\documentclass[dvipdfmx]{article}
\usepackage{tikz}
\usepackage{amssymb}
\usepackage{graphicx}

\usepackage{amssymb}
\usepackage{graphicx}
\newtheorem{teiri}{Theorem}

\newtheorem{kei}{Corollary}
\newtheorem{rei}{Example}

\newtheorem{teigi}{Definition}
\def\ci{\perp\!\!\!\perp}

\begin{document}

\title{
A Theoretical Analysis of the BDeu Scores in Bayesian Network Structure Learning
}
\author{
Joe Suzuki\\Osaka University
}
\date{}
\maketitle

In Bayes score-based Bayesian network structure learning (BNSL), 
we are to specify two prior probabilities:
over the structures and over the parameters.
In this paper, we mainly consider the parameter priors, in particular for the BDeu (Bayesian Dirichlet
equivalent uniform) and Jeffreys' prior.
In model selection, given examples, we typically consider how well a model explains the examples and 
how simple the model is, and choose the best one for the criteria.
In this sense, if a model A is better than another model B for both of the two criteria, 
it is reasonable to choose the model $A$. In this paper, we prove that the BDeu violates 
such a regularity,  and that we will face a fatal situation in BNSL:
the BDeu tends to add a variable to the current parent set of a variable $X$ even when 
the conditional entropy reaches to zero. In general, priors should be reflected by  the learner's belief, and should not be rejected 
from a general point of view.  However, this paper suggests that the underlying belief of the BDeu contradicts with our intuition
in some cases, which 
has not been known until this paper appears.


\section{Introduction}

In this paper, we consider learning a Bayesian network (BN) structure from examples,
where a BN \cite{pearl} is defined to be a directed acyclic graph (DAG) that expresses factorization of the distribution.
In particular, we find a structure with the maximum posterior probability given examples
w.r.t. the  prior probabilities over parameters and structures.

In order to  compare the posterior probabilities of the BN structures with $N$ variables, 
we construct quantities (scores) for each of the $2^N$ subsets. 
For example, if there are $N=3$ variables $X, Y, Z$, then we compute from $n$ triples of examples 
eight  scores w.r.t. $\{\}$, $\{X\}$, $\{Y\}$, $\{Z\}$, $\{Y,Z\}$, $\{Z,X\}$, $\{X,Y\}$, and $\{X,Y,Z\}$.
The prior probability over structures is usually the uniform distribution, and
the posterior probability of each structure can be computed from those scores.
In BN structure learning (BNSL), the Dirichlet distribution
is being used for expressing the prior probability over the parameters. For example, for variable $X$, the prior probabilities over the parameters $\theta=(\theta_x)$
is expressed as $\prod_x\theta_x^{a(x)-1}$ multiplied by a constant, where $\theta_x$ is 
the probability of $X=x$ and $a(x)$ is a positive constant associated with $X=x$.
In this sense,  the choice of constant $a(\cdot)$  determines the solution of BNSL.

Our goal of BNSL in this paper is to seek a structure with the maximum posterior probability.
We assume that a true generating probability exists that is expressed by a structure and parameters,
and they should be estimated as correctly as possible.
When the sample size $n$ is small, the obtained structure may not be correct.
However, combining the prior probability based on learner's belief and the data of size $n$,
we obtain the best solution. In this paper, we discuss the choice of constants $a(\cdot)$ and its resulting BNSL.

Several settings of $a(\cdot)$ have been  used for BNSL: the BD scores based on Jeffreys' prior \cite{jeffreys} ($a(\cdot)=0.5$),
the BDe scores ($a(x)$ is proportional to the probability of $X=x$) \cite{heck}, and 
the BDeu (BD equivalent uniform) scores ($a(x)=\delta/\alpha$ with $\alpha$ the number of 
values that $X$ takes and $\delta>0$ a constant called an equivalent sample size) \cite{wray} 

Among them, the BDeu scores are used often in BNSL, and its performance has been examined most.
Steck and Jaakkola \cite{sj} demonstrated that as the
equivalent sample size $\delta$ and the sample size $n$ decreases and increases, respectively,
simpler BN structures tend to be selected.
Silander, Kontkanen, and Myllymaki \cite{skm} performed
empirical experiments to find the optimum equivalent sample size $\delta$. 
They confirmed the result by Steck and Jaakkola \cite{sj} and indicated
that the solution is highly sensitive to the chosen $\delta$. 
Although the analysis on BDeu was only empirically, 
Ueno \cite{uenob} provided an asymptotic
analysis of the scores
for BNSL, and suggested 
the ratio of $\delta>0$ and sample size $n$ determines
the penalty of adding arcs in BNSL. 

The original paper by 
Heckerman et. al. \cite{heck} compared variants of BDe (including BDeu) and K2 \cite{cf} based on
 Jeffereys' prior in this paper
to conclude that "scores using relatively uninformative priors such as Jeffreys' performed as well as or better than scores
using more informative priors such as BDeu, unless that prior knowledge was extremely accurate".
The results seems to be reasonable but does not give information on any merits the BDeu in the general setting.

The main drawback of the existing research on the scores is not to answer
why the BDeu scores should be used rather than the BD scores based on Jeffreys' prior
(they mainly consider what $\delta>0$ should be chosen when they use the BDeu scores).
The evaluations of the existing researches were mainly experimental and 
little persuasive results have been obtained.

It is not hard to prove that BNSL based on the BDeu scores obtains a correct structure as the sample size $n$ grows.
But, thus far, nobody noticed that such consistency is true in the sense of pointwise property (not uniform).
This means that for the BDeu, 
for any large $n$, there may exist a BN such that some conditional independence (CI) relations are
not detected  while for each fixed BN, all the CI relations are detected for large $n$.

In BNSL, we seek the parent set $U$ of each variable $X$ that maximizes the 
the conditional score of $X$ given $U$.
Then, we require balance between 
the fitness of $U$  to the $n$ examples and the size of $U$.
If the conditional entropy of $X$ given $U$ is no more than that given $U'$
and $U$ is contained in $U'$, then the conditional score of parent set $U$ should be no less than that of parent set $U'$.
It is straightforward to justify such a regularity in model selection: if a model A is better than B w.r.t. both of
fitness and simplicity, then model A should be chosen.

We prove that the BDeu scores violate regularity, and suggest that this phenomenon causes serious problems in BNSL.
Although Silander, Kontkanen, and Myllymaki \cite{skm} pointed out that the balance is not valid in the BDeu scores,
but they did not prove any  mathematical statement.
On the other hand, the BD scores based on Jefferys' prior do not face such a problem, which we find in this paper is 
due to its uniform consistency in BNSL.

For example, any scientist considers balance between simplicity of a hypothesis and fitness of experimental data to the hypothesis.
Even if the hypothesis explains the data well, if it was complicated, the scientist would not accept it as
the law that generated the data. Scientists generally prefer simple expressions such as  Newton's laws of motion, Maxwell's equations.
However, the BDeu might choose the more complicated one even if two hypotheses explained data equally well (this paper
assumes a strict Occam's razor point of view regarding model selection, and that all "fatality statements pertain to this assumption").

One might think that each BDeu prior over parameters may represent his/her prior belief and cannot be rejected from a general point of view. 
However, we suggest that their priors might have been based on inappropriate beliefs that support 
irregularity: if they knew that using the BDeu causes such situations, their beliefs and the corresponding prior 
might change in the future.

This paper is organized as follows. Section 2 gives preliminaries: Sections 2.1, 2.2, 2.3, 2.4, and 2.5 
explain BN, scores, BNSL, BDeu and Jeffreys, and consistency, respectively. Section 3 proves new results (Theorems 1, 2, 3):
Theorem 1 in Section 3.1 suggests that the consistency of the BDeu is only pointwise;
Theorem 2 in Section 3.2 gives a general statement that explains properties of the BDeu; and Theorem 3 in Section 3.3 shows that when 
the BDeu scores face fatal situations in BNSL. Section 4 summaries the properties that this paper proves and their significances
and states a future work.

\section{Background}
In this section, we give basic materials to understand the results in the next section.

\subsection{Bayesian Network}

In this paper, we define a Bayesian network (BN) to be a directed acyclic graph (DAG) that expresses factorization 
of the distribution $P(X^{(1)},\cdots,X^{(N)})$,
where $X^{(1)},\cdots,X^{(N)}$ are random variables that take finite numbers of values ($N\geq 1$).
For example, for $N=3$ variables $X$, $Y$, $Z$, we 
can check that the 25 factorizations such as $P(X)P(Y|X)P(Z|X,Y)$ and $P(X)P(Y|X)P(Z|X)$
are categorized into the eleven equivalent classes:
$$P(X)P(Y)P(Z), P(X)P(Y,Z),P(Y)P(Z,X),P(Z)P(X,Y)$$
$$\frac{P(ZX)P(X,Y)}{P(X)}\ ,\ \frac{P(X,Y)P(Y,Z)}{P(Y)}\ ,\ \frac{P(ZX)P(X,Y)}{P(Z)}$$
$$\frac{P(Y)P(Z)P(X,Y,Z)}{P(Y,Z)}\ ,\ \frac{P(Z)P(X)P(X,Y,Z)}{P(Z,X)}\ ,\ \frac{P(X)P(Y)P(X,Y,Z)}{P(X,Y)}\ ,\ {\rm and}\ P(X,Y,Z)$$

\subsection{Bayesian Dirichlet Scores}
In this subsection, we consider to give a score to each sequence of length $n$.

In general, there are $\alpha^n$ possible sequences for the $n$ realizations of random variable $X$
that takes $\alpha$ values because each realization is one of the $\alpha$ elements.

Suppose that we do not know the probability $\theta_x$ of $X=x$ and we wish to assign a probability to
 each sequence of length $n$.
 One way to deal with this problem is to prepare a weight $w(\cdot)$ over $\theta=(\theta_x)$
 such as
$$w(\theta)=K\prod_x\theta_x^{a(x)-1}\ $$
with $\displaystyle K=[\int \prod_x\theta_x^{a(x)-1}d\theta]^{-1}$ to obtain the quantity (Bayesian Dirichlet score)
\begin{eqnarray}\label{eq71}
Q^n(X)&=&\int \prod_{x} \theta^{c(x)}w(\theta)d\theta=K\int \prod_{x} \theta^{c(x)+a(x)-1}d\theta\nonumber\\
&=&\frac{\Gamma(\sum_x a(x))}{\Gamma(n+\sum_x a(x))}\prod_x \frac{\Gamma(c(x)+a(x))}{\Gamma(a(x))}
\ ,
\end{eqnarray}
where $a(x)>0$ is a constant that may depend on $X=x$, 
$c(x)$ is the frequencies of $X=x$ in the sequence, and $\Gamma(\cdot)$ is the Gamma function
$\Gamma(z)$ that generalizes factorial $n!=n(n-1)\cdots 1$, if $z=n+1$ for an integer $n$.
In particular, we use the following property:
$$\frac{\Gamma(n+b)}{\Gamma(b)}=(n-1+b)(n-2+b)\cdots b$$
for integer $n\geq 0$ and real $b>0$. For example, in (\ref{eq71}), 
$$\frac{\Gamma(c(x)+a(x))}{\Gamma(a(x))}=(c(x)-1+a(x))(c(x)-2+a(x))\cdots a(x)$$
for each $x$, and
$$\frac{\Gamma(n+\sum_xa(x))}{\Gamma(\sum_xa(x))}=(n-1+\sum_xa(x))(n-2+\sum_xa(x))\cdots \sum_x a(x)\ .$$

Similarly,  we express $Q^n(X,Y)$ using constants $a(x,y)>0$ and frequencies $c(x,y)$ by
\begin{eqnarray}\label{eq711}
Q^n(X,Y)=\frac{\Gamma(\sum_{x}\sum_y a(x,y))}{\Gamma(n+\sum_{x}\sum_y a(x,y))}\prod_{x}\prod_y
\frac{\Gamma(c(x,y)+a(x,y))}{\Gamma(a(x,y))}\ .
\end{eqnarray}
In general, we express $Q^n(S)$ with $S\subseteq V$ using $a(s)>0$ and $c(s)$
\begin{eqnarray*}\label{eq712}
Q^n(S)=\frac{\Gamma(\sum_sa(s))}{\Gamma(n+\sum_sa(s))} \prod_{s}\frac{\Gamma(c(s)+a(s))}{\Gamma(a(s))}\ ,
\end{eqnarray*}
where $S$ takes $\gamma$ values and $s$ ranges over the $\gamma$ values.
For example,  if $X, Y\in V$ take $\alpha$ and $\beta$ values, then 
$S=\{X,Y\}$ takes $\gamma=\alpha\beta$ values.

\subsection{Bayesian Network Structure Learning}
In this subsection, 
based on the idea to give a score to each sequence of length $n$, we consider to construct a procedure of BNSL.

Suppose we wish to test whether random variables $X$ and $Y$ are independent\footnote{We denote $X\ci Y|Z$ if $X$ and $Y$ are conditionally independent given $Z$}
from $n$ pairs of examples $x^n=(x_1,\cdots,x_n)$ and $y^n=(y_1,\cdots,y_n)$ following $X$ and $Y$, respectively.
If we take the Bayes score-based approach, 
we prepare the prior probability $0<p<1$ of $X\ci Y$ and the scores $Q^n(X),Q^n(Y),Q^n(X,Y)$ of 
$x^n,y^n,(x^n,y^n)$, and  decide that $X\ci Y$ \cite{suzuki12} if and only if 
\begin{equation}\label{eq19}
pQ^n(X)Q^n(Y)\geq (1-p)Q^n(X,Y)\ ,
\end{equation}
where we assume that the constants $a(x)$, $a(y)$, and $a(x,y)$ in $Q^n(X),Q^n(Y),Q^n(X,Y)$ have been 
specified beforehand.
The decision (\ref{eq19}) maximizes the posterior probability of either $X\ci Y$ or $X\not \ci Y$
given the prior probability $p$, and satisfies consistency (see Section 2.5).

Similarly, 
suppose we wish to test whether $X$ and $Y$ are
conditionally  independent given another random variable $Z$
from $n$ triples of examples 
$x^n=(x_1,\cdots,x_n)$, $y^n=(y_1,\cdots,y_n)$, $z^n=(z_1,\cdots,z_n)$
following $X$, $Y$ and $Z$, respectively.
If we take the Bayes score-based approach, 
we prepare the prior probability $0<p<1$ of $X\ci Y|Z$ and
the scores  $Q^n(X,Z),Q^n(Y,Z),Q^n(Z),Q^n(X,Y,Z)$ of 
$(x^n,z^n),(y^n,z^n),z^n,(x^n,y^n,z^n)$, and decide that $X\ci Y|Z$ \cite{suzuki15} if and only if 
\begin{equation}\label{eq192}
pQ^n(X,Z)Q^n(Y,Z)\geq (1-p)Q^n(X,Y,Z)Q^n(Z)
\end{equation}

In this paper, we consider to learn the BN structure given $n$ i.i.d. tuples of examples 
$$\{(X^{(1)},\cdots,X^{(N)})=(x_{i,1},\cdots,x_{i,N})\}_{i=1}^n$$
w.r.t. $N$ variables
$X^{(1)},\cdots,X^{(N)}$,
where
we assume 
\begin{enumerate}
\item no missing values in the $n$ tuples of examples, and 
\item the prior probabilities of the structures are given. 
\end{enumerate}
For example, if $N=3$,  the problem is to choose one of the eleven factorizations, and we compare the values
{\small
$$Q^n(X)Q^n(Y)Q^n(Z), Q^n(X)Q^n(Y,Z),Q^n(Y)Q^n(Z,X),Q^n(Z)Q^n(X,Y)$$
$$\frac{Q^n(Z,X)Q^n(X,Y)}{Q^n(X)}\ ,\ \frac{Q^n(X,Y)Q^n(Y,Z)}
{Q^n(Y)}\ ,\ \frac{Q^n(Z,X)Q^n(X,Y)}{Q^n(Z)}$$
$$\frac{Q^n(Y)Q^n(Z)Q^n(X,Y,Z)}{Q^n(Y,Z)}\ ,\ \frac{Q^n(Z)Q^n(X)Q^n(X,Y,Z)}{Q^n(Z,X)}\ ,\ \frac{Q^n(X)Q^n(Y)
Q^n(X,Y,Z)}{Q^n(X,Y)}\ ,\ Q^n(X,Y,Z)$$}
multiplied by their prior probabilities, and obtain 
a BN structure of the maximum posterior probability \cite{cf}.

\subsection{Jeffreys' Prior and BDeu}
There are many ways to give constants $a(s)$ introduced in Section 2.2 for each state $s$.

In this paper, we consider two settings of $a(s)$:
\begin{enumerate}
\item $a(s)=0.5$ for the $\gamma$ values and for all $S\subseteq V$; and
\item $a(s)=\delta/\gamma$  for  the $\gamma$ values, where $\delta$ is a constant (equivalent sample size)  shared by all $S\subseteq V$.
\end{enumerate}
We say that the values $Q^n(\cdot)$ based on them are the BD (Bayesian Dirichlet) scores based on Jeffreys' prior \cite{jeffreys} and 
BDeu (BD equivalent uniform \cite{wray} \cite{uenob}) score, respectively.

The idea of using the BDeu scores is based on the following: 
in order to obtain $Q^n(Y|X)$ in the form 
\begin{equation}\label{eq155}
Q^n(Y|X)=\prod_x\{
\frac{\Gamma(a(x))}{\Gamma(c(x)+a(x))}\prod_{y}\frac{\Gamma(c(x,y)+a(x,y))}{\Gamma(a(x,y))}\}\ ,
\end{equation}
the terms
\begin{equation}\label{eq157}
\frac{\Gamma(\sum_x a(x))}{\Gamma(n+\sum_x a(x))}\ {\rm and}\ 
\frac{\Gamma(\sum_{x}\sum_y a(x,y))}{\Gamma(n+\sum_{x}\sum_y a(x,y))}
\end{equation}
in (\ref{eq71}) and (\ref{eq711}) should be cancelled out, so that
we require $a(x)=\sum_y a(x,y)$.
If we set $\delta=\sum_xa(x)$ and assume that $a(x)$ and $a(x,y)$ are constant for the $\alpha$ and $\alpha\beta$ values, respectively, 
then $a(x)=\delta/\alpha$ and $a(x,y)=\delta/(\alpha\beta)$ follow, so that we obtain conditional score\footnote{
If $Y=X^{(i)}$ and $X$ is a parent set of $Y$ in $Q^n(Y|X)$, the quantity $c(x,y)$ and $a(x,y)$ are expressed by $N_{ijk}$ and $N'_{ijk}$,
respectively, when $x=j$ and $y=k$ in the literature. For ease of understanding, however, in this subsection, we use the current notation.}
$$Q^n(Y|X)=\prod_x\{
\frac{\Gamma(\delta/\alpha)}{\Gamma(c(x)+\delta/\alpha)} 
\prod_{y}\frac{\Gamma(c(x,y)+\frac{\delta}{\alpha\beta})}{\Gamma(\frac{\delta}{\alpha\beta})}\}\ ,\ $$
and score
$$Q^n(X,Y)=
\frac{\Gamma(\delta)}{\Gamma(n+\delta)} \prod_{x}\prod_y\frac{\Gamma(c(x,y)+\frac{\delta}{\alpha\beta})}{\Gamma(\frac{\delta}{\alpha\beta})}\ ,$$
which is  generalized to 
$$Q^n(S)=\frac{\Gamma(\delta)}{\Gamma(n+\delta)} \prod_{s}\frac{\Gamma(c(s)+\frac{\delta}{\gamma})}{\Gamma(\frac{\delta}{\gamma})}\ .$$
when $\gamma$ is the number of values that $S$ takes and $s$ ranges over the $\gamma$ values.

Note that if we plug-in $a(x,y)=0.5$ into (\ref{eq71})(\ref{eq711})(\ref{eq155}),
then $Q^n(X,Y)$ and $Q^n(X)Q^n(Y|X)$ do not coincide while the two coincide for the BDeu \cite{k-f}.
In particular, if we assume the form in (\ref{eq155}), $Q^n(X,Y)=Q^n(X)Q^n(Y|X)$
 (score equivalence) if and only if the  score is BDeu (Theorem 18.4 \cite{k-f}).
However, in this paper, we redefine the conditional score $Q^n(Y|X)$ by
$Q^n(X,Y)/Q^n(X)$ without using (\ref{eq155}),
which insures that all the conditional scores based on the unconditional scores (\ref{eq71}) satisfy score equivalence even if $a(x,y)=0.5$.

\subsection{Consistency of Learning Bayesian Network Structures}

In any estimation of statistics, we expect that 
the estimated value approaches to the correct value as the sample size grows.

We say that a model selection procedure is (strongly) consistent if 
the probability one is assigned to the sequences of length $n$ such that
the estimated model is correct only but finite times as $n\rightarrow \infty$ \cite{billingsley}.

In particular, for
BNSL,
we require
$Q^n(\cdot)$ to satisfy that for any three disjoint subsets
${\mathbf X}$, ${\mathbf Y}$, and ${\mathbf Z}$ of $V$,
the following decision should be consistent: 
\begin{equation}\label{eq69}
{\mathbf X}\ci {\mathbf Y}|{\mathbf Z} \Longleftrightarrow
{Q^n({\mathbf X}\cup {\mathbf Z})Q^n({\mathbf Y}\cup {\mathbf Z})}\geq {Q^n({\mathbf X}\cup{\mathbf Y}\cup{\mathbf Z})Q^n({\mathbf Z})}\ .
\end{equation}
For consistency, the reference  \cite{suzuki12} showed under $a(x)=a(y)=a(x,y)=0.5$ in (\ref{eq711})
$$(\ref{eq19})\Longleftrightarrow X\ci Y$$
and
$$\frac{1}{n}\log \frac{(1-p)Q^n(X,Y)}{pQ^n(X)Q^n(Y)}\rightarrow I(X,Y)$$
for large $n$ with probability one as $n\rightarrow \infty$,
where $I(X,Y)$ is the mutual information between $X,Y$ which is defined by
$$I(X,Y):=\sum_x\sum_y P_{XY}(x,y)\log \frac{P_{XY}(x,y)}{P_{X}(x)P_{Y}(y)}\ .$$
Since $I(X,Y)$ is nonnegative, the estimator of $I(X,Y)$ converges to zero as $n\rightarrow \infty$ if and only if $X\ci Y$.
In fact, the first statement can be proved in many ways. For example, 
we can describe $x^n$ and $y^n$ separately in 
$$-\log p-\log Q^n(X)-\log Q^n(Y)$$
bits (if the logarithm base is two) assuming $X\ci Y$
while we can describe them together in 
$$-\log (1-p)-\log Q^n(X,Y)$$
bits assuming $X\not\ci Y$
because the binary information $X\ci Y$ and $X\not\ci Y$ can be described in $-\log p$ and $-\log (1-p)$ bits,
respectively, and $Q^n(X),Q^n(Y),Q^n(X,Y)$ satisfy Kraft's inequalities \cite{mdl}:
$$\sum_{x^n}Q^n(X)\leq 1\ ,\ \sum_{y^n}Q^n(Y)\leq 1\ ,\ \sum_{x^n}\sum_{y^n}Q^n(X,Y)\leq 1\ .$$
Strong consistency of the minimum description length (MDL) principle \cite{mdl} implies the claim.
For other derivation, see \cite{suzuki12}.

On the other hand, the reference \cite{suzuki15} showed under $a(x,y,z)=a(x,z)=a(y,z)=a(z)=0.5$,
\begin{equation}\label{eq201}
(\ref{eq192})\Longleftrightarrow X\ci Y|Z
\end{equation}
and
$$\frac{1}{n}\log \frac{(1-p)Q^n(X,Y,Z)Q^n(Z)}{pQ^n(X,Z)Q^n(Y,Z)}\rightarrow I(X,Y|Z)$$
for large $n$ with probability one as $n\rightarrow \infty$,
where $I(X,Y|Z)$ is the conditional mutual information between $X,Y$ given $Z$ which is defined by
$$I(X,Y|Z):=\sum_x\sum_y\sum_z P_{XYZ}(x,y,z)\log \frac{P_{XYZ}(x,y,z)P_Z(z)}{P_{XZ}(x,z)P_{YZ}(y,z)}\ .$$
Since $I(X,Y|Z)$ is nonnegative, the estimator of $I(X,Y|Z)$ converges to zero as $n\rightarrow \infty$ if and only if $X\ci Y|Z$.

The straightforward extension of (\ref{eq201}) such that $X\rightarrow {\mathbf X}$, $Y\rightarrow {\mathbf Y}$, and $Z\rightarrow {\mathbf Z}$
implies the condition (\ref{eq69}) for the BD score based on Jeffreys' prior
because $I(X,Y|Z)=0 \Longleftrightarrow X\ci Y|Z$.



\section{Properties of BDeu}

In this section, we derive properties of the BDeu scores that have not been found thus far.

Let ${\mathbf X},{\mathbf Y},{\mathbf Z}$ be disjoint subsets of $V=\{X^{(1)},\cdots,X^{(N)}\}$, and 
$\alpha,\beta,\gamma$ the numbers of values that ${\mathbf X},{\mathbf Y},{\mathbf Z}$ take.
Given $n$ examples of ${\mathbf X},{\mathbf Y},{\mathbf Z}$, we estimate whether ${\mathbf X}$ and ${\mathbf Y}$ are 
independent given ${\mathbf Z}$ based on 
the quantity
\begin{equation}
J(n):=\frac{1}{n}\{\log {Q^n({\mathbf X},{\mathbf Y},{\mathbf Z})}+\log {Q^n({\mathbf Z})}-\log {Q^n({\mathbf X},{\mathbf Z})}
-\log {Q^n({\mathbf Y},{\mathbf Z})}\}\ .
\end{equation}
Let $J_*(n)$ and $J^*(n, \delta)$ be the values of $J(n)$ for the BD score based on Jeffreys' prior and the BDeu score 
with equivalent sample size $\delta>0$, respectively.

\subsection{Consistency of BDeu is pointwise}

It is well known that the Bayesian scores are consistent (Corollary 18.2 \cite{k-f}).
In fact, (\ref{eq69}) holds for large $n$, which means both $J_*(n)$ and $J^*(n,\delta)$ diminishes to zero for large $n$
if and only if ${\mathbf X}\ci {\mathbf Y}|{\mathbf Z}$.

However, what we show in this subsection is more specific. Let $\theta$ be the parameters that control ${\mathbf X}$, ${\mathbf Y}$, and ${\mathbf Z}$.
We prove that $\sup_\theta J^*(n,\delta)$ remains away from zero for any $\delta>0$ even when $n$ grows while 
$\sup_\theta J_*(n)$ diminishes.


Let 
\begin{equation}\label{eq9.5}
I(n):=\sum_x\sum_y\frac{c(x,y,z)}{n}\log\frac{c(x,y,z)c(z)}{c(x,z)c(y,z)}-\frac{(\alpha-1)(\beta-1)\gamma}{2n}\log {n}\ ,
\end{equation}
where $x,y$, and $z$ range over the values of ${\mathbf X},{\mathbf Y}$, and ${\mathbf Z}$.
Then, we obtain the following theorem. For the proof, see Appendix A.
\begin{teiri}\rm 
\begin{equation}\label{eq412}
J_*(n)=I(n)+O(\frac{1}{n})
\end{equation}
and
\begin{eqnarray}\label{eq112}
J^*(n,\delta)&=&I(n)+\frac{1}{n}D_n+O(\frac{1}{n})
\end{eqnarray}
with
\begin{eqnarray}\label{eq145}
D_n&=&-(\frac{\delta}{\alpha\gamma}-\frac{1}{2})\sum_{x}\sum_z\log \frac{c(x,z)+\delta/\alpha\gamma}{n+\delta}
-(\frac{\delta}{\beta\gamma}-\frac{1}{2})\sum_{y}\sum_z\log \frac{c(y,z)+\delta/\beta\gamma}{n+\delta}\nonumber\\&&+
(\frac{\delta}{\alpha\beta\gamma}-\frac{1}{2})\sum_x\sum_y\sum_z\log \frac{c(x,y,z)+\delta/\alpha\beta\gamma}{n+\delta}
+(\frac{\delta}{\gamma}-\frac{1}{2})\sum_z\log \frac{c(z)+\delta/\gamma}{n+\delta}\ ,\nonumber\\
\label{eq413}
\end{eqnarray}
where  $f=g+O(h)$ denotes that $(f-g)/h$ is bounded by constants that depend on at most $\alpha$, $\beta$, $\gamma$, $\delta$.
\end{teiri}

We claim that consistency of the BDeu is pointwise in the following sense.
\begin{rei}\label{rei11}\rm
Suppose that 
${\mathbf X}=\{X\}$,
${\mathbf Y}=\{Y\}$,
${\mathbf Z}=\{\}$,
$\alpha=\beta=2$, $\gamma=1$, $\delta=1$, and that the logarithm base in (\ref{eq145}) is two. Then,
$$D_n=-\frac{1}{4}\sum_{x=0}^1\sum_{y=0}^1\log \frac{c(x,y)+1/4}{n+1}\ ,$$
and the last terms of $I(n)$ is $\displaystyle \frac{1}{2n}\log n$ because $\alpha=\beta=2$ and $\gamma=1$.
First of all, $D_n/n$ converges to zero uniformly for any $c(\cdot,\cdot)$
because each of the four terms is at most $\displaystyle \frac{1}{4}\log(n+1)+\frac{1}{2}$, thus 
$\displaystyle \frac{D_n}{n}\leq \frac{\log(n+1)+2}{n}$.
However, if the probabilities of $X=1$ and $Y=1$ are too small, the sum of the three terms divided by $n$
w.r.t. $(x,y)=(1,1),(1,0),(0,1)$ may exceed $\displaystyle \frac{1}{2n}\log n$,
which means that $I(n)+\frac{1}{n}D_n$ is positive even if $X\ci Y$, thus $J^*(n,\delta)>0$. 
On the other hand, we know $J_*(n)\leq 0\Longleftrightarrow X\ci Y$ for large $n$ (see Section 2.5).
\end{rei}

\begin{figure}\label{fig1}
\input{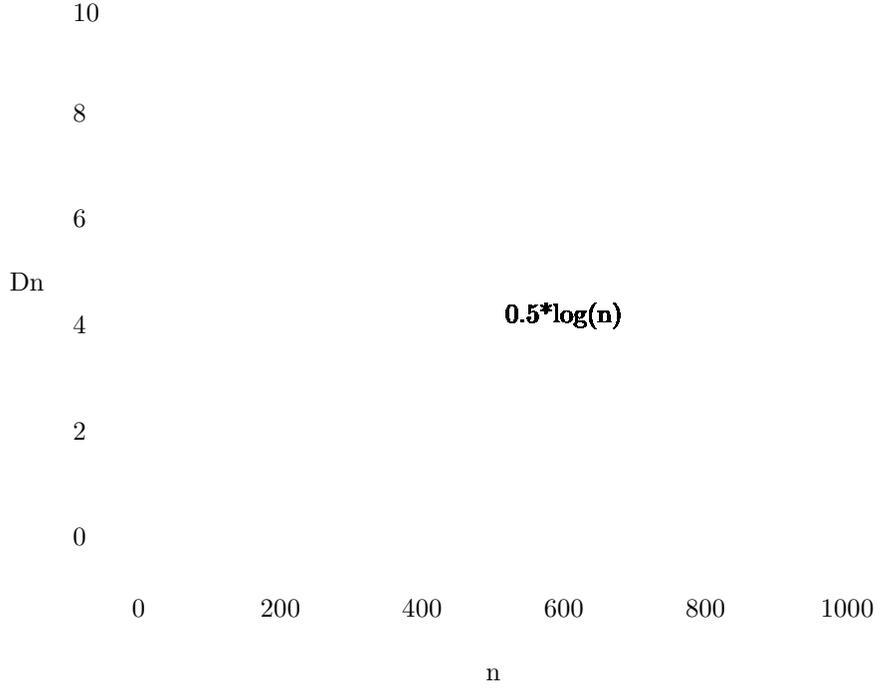}
\caption{Randomly Generated Sequence $D_n$: we observe that the plots of $D_n$ are above the curve ($0.5 \log n$) for all $10\leq n\leq 1000$.}
\end{figure}

Let $\theta$ be the parameters that control ${\mathbf X}$,
${\mathbf Y}$, ${\mathbf Z}$.
such as $P(X=1)$ and $P(Y=1)$. We observe that when ${\mathbf X}\ci {\mathbf Y}|{\mathbf Z}$,
$$\lim_{n\rightarrow \infty}\sup_\theta P_\theta\{J_*(n)\leq 0 \}=1\ ,$$
$$\lim_{n\rightarrow \infty}P_\theta\{J^*(n,\delta)\leq 0 \}=1$$
for each $\theta$, and 
$$\lim_{n\rightarrow \infty}\sup_\theta P_\theta\{J^*(n,\delta)\leq 0 \}<1\ .$$
In fact, in Example \ref{rei11}, if we set $P(X=1)=P(Y=1)=n^{-0.75}$ and $X\ci Y$, then the values of $D_n$
are distributed as in Figure \ref{fig1}.
We observe that all the plot are above $0.5\log n$ whereas they have been  randomly generated.
This means that $J^*(n,\delta)>0$ with an extremely high probability when 
$P(X=1)=P(Y=1)=n^{-0.75}$, even if  $X\ci Y$.
In this sense, we say that the consistency of the BDeu is pointwise while 
that of Jeffreys' is uniform.
For the definitions of poinwise and uniform convergences in a general sense, consult the reference \cite{rudin}.

It seems that the problem is trivial because 
it is rare that $P(X=1)$ and $P(Y=1)$ are so small. 
However, in the later subsections,
we prove a general statement (Section 3.2) that contains Example \ref{rei11} 
and show that it causes  a fatal problem in BNSL (Section 3.3).



\subsection{A Property of BDeu implied by Pointwise Consistency}

In this subsection, we consider what problem arises from pointwise consistency,
In Example \ref{rei11}, If we observe the sequences of $n$ zeros for $X$ and $Y$ without knowing the values of $P(X=1)$ and $P(Y=1)$,
it is likely that $X$ and $Y$ are independent because $P(X=0)=P(Y=0)=P(X=Y=0)=1$ implies
$P(X)P(Y)=P(XY)$.  Then, we expect (\ref{eq19}), in particular $Q^n(X)Q^n(Y)\geq Q^n(XY)$
if we assume the prior probability $p$ of $X\ci Y$ to be 0.5.
However, we observe 
$Q^n(X)Q^n(Y) < Q^n(X,Y)$
for the BDeu score with any  $\delta>0$
while
$Q^n(X)Q^n(Y)\geq  Q^n(XY)$
for the BD score based on Jeffreys' prior. For example, suppose $n=5$ and $\delta=1$.
$$Q^n(X)=Q^n(Y)=\frac{\Gamma(1)}{\Gamma(n+1)}\cdot \frac{\Gamma(n+1/2)}{\Gamma(1/2)}=\frac{1}{5!}\cdot \frac{9}{2} \cdot \frac{7}{2}\cdot \frac{5}{2}\cdot \frac{3}{2} \cdot \frac{1}{2}=
0.246\cdots\ ,$$
thus $Q^n(X)Q^n(Y)=0.0605 \cdots$ for the both scores. On the other hand, 
$$Q^n(X,Y)=\frac{\Gamma(1)}{\Gamma(n+1)}\cdot \frac{n+1/4}{\Gamma(1/4)}=\frac{1}{5!}\cdot\frac{17}{4}\cdot\frac{13}{4}\cdot\frac{9}{4}
\cdot\frac{5}{4}\cdot\frac{1}{4}=0.0809\cdots$$
for BDeu, and
$$Q^n(X,Y)=\frac{\Gamma(2)}{\Gamma(n+2)}\cdot \frac{n+1/2}{\Gamma(1/2)}=\frac{1}{6!}\cdot\frac{9}{2}\cdot\frac{7}{2}\cdot\frac{5}{2}
\cdot\frac{3}{2}\cdot\frac{1}{2}=0.0410\cdots$$
for Jeffreys'.

In fact, by extending the example, 
we obtain the main theorem of this paper as follows
\begin{teiri}\rm 
Suppose we have examples $(x_1,\cdots,x_n)$, $(y_1,\cdots,y_n)$, and $(z_1,\cdots,z_n)$
that follow ${\mathbf X}$,${\mathbf Y}$, and ${\mathbf Z}$, respectively. If functions $f$ and $g$ exist such that $x_i=f(z_i)$ and $y_i=g(z_i)$
for $i=1,\cdots,n$ and we construct scores $Q^n(\cdot)$ based on the examples for $n\geq 1$, then 
$J_*(n)\leq 0$ and 
$J^*(n,\delta)\geq 0$ for any $\delta>0$.
In particular, $J^*(n,\delta)>0$ for any $\delta>0$ and $n\geq 2$
\end{teiri}
See Appendix B for the proof.

The following corollary gives a solution of the aforementioned example.
\begin{kei}\rm 
Suppose we have examples $(x_1,\cdots,x_n)$ and $(y_1,\cdots,y_n)$
that follow $X$ and $Y$, respectively. If $x_1=\cdots=x_n$ and $y_1=\cdots=y_n$ for $n\geq 2$, and 
we construct scores $Q^n(\cdot)$ based on the examples, then 
$Q^n(X)Q^n(Y)\geq  Q^n(X,Y)$
for the score based on Jeffreys' prior and 
$Q^n(X)Q^n(Y) < Q^n(X,Y)$
for the BDeu score with any $\delta>0$.
\end{kei}

\begin{figure}
\begin{center}
\input{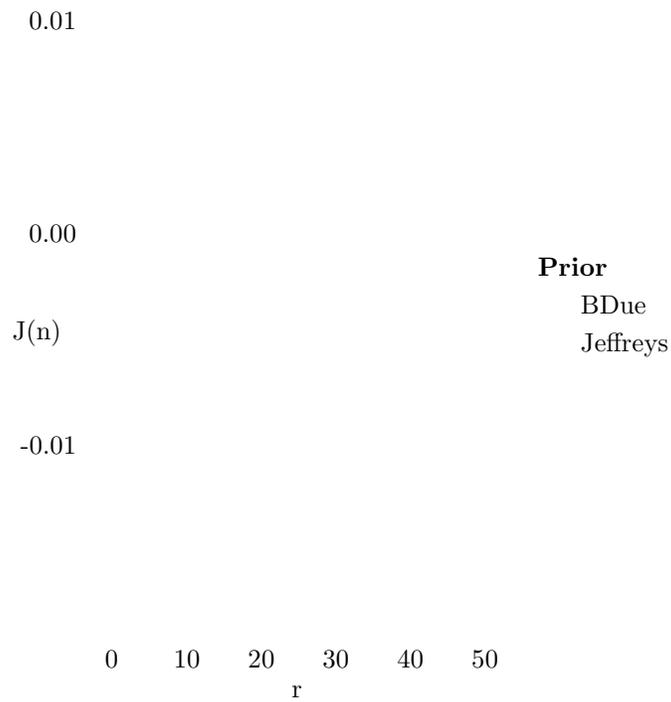}
\end{center}
\caption{
The values of $J(n)$ for the BDeu scores and the BD scores based on Jeffreys' prior: the former is positive for $r=0,1,2,3$
while the latter keeps the same value for all $r$.
\label{fig126}}
\end{figure}

Note that the phenomenon is due to singularity at specific cases such as $n$ zeros.
For example, $Q^n(X)Q^n(Y)< Q^n(X,Y)$ for BDeu occurs  often even when $P(X=0,Y=0)\not=1$. Figure \ref{fig126} shows the value of $J(n)$
when the sequence of $X$ contains $r$ ones with $0\leq r\leq n/2$ and that of $Y$ is $n=100$ zeros. 
The $J(n)$ value is positive w.r.t. $r=0,1,2,3$ for the BDeu scores while  
it is always negative and keeps the same value 
\begin{eqnarray*}
&&\frac{1}{n}\log Q^n(X,Y)-\frac{1}{n}\log Q^n(X)-\frac{1}{n}\log Q^n(Y)\\
&=&\frac{1}{n}\log \{\frac{\Gamma(r+\frac{1}{2})\Gamma(n-r+\frac{1}{2})\Gamma(\frac{1}{2})^2}{\Gamma(n+2)}\}
-\frac{1}{n}\log \{\frac{\Gamma(r+\frac{1}{2})\Gamma(n-r+\frac{1}{2})}{\Gamma(n+1)}\}\\
&&-\frac{1}{n}\log \{\frac{\Gamma(n+\frac{1}{2})\Gamma(\frac{1}{2})}{\Gamma(n+1)}\}\\
&=&\frac{1}{n}\log \frac{\sqrt{\pi}\Gamma(n+1)}{(n+1)\Gamma(n+\frac{1}{2})}
\end{eqnarray*}
w.r.t.  $0\leq r\leq n$ for the BD scores based on Jeffreys' prior. Note that $X$ and $Y$ are independent
if we assume $P(Y=0)=1$ regardless of the value of $X$.

It does not seem that anything inconvenient occurs due to Corollaries 1 for the case
because it is unlikely that zeros continue so many times even if a couple of ones (Figure \ref{fig126}) may be contained in one of $X$ and $Y$.
In this sense, Corollary 1 may be only of theoretical interest and may not be useful.
However, Theorem 2 that generalizes Corollary 1 contains cases such that we face fatal situations, as we will see in the next subsection.

\subsection{What happens when BDeu is applied to BNSL}

In this subsection, we consider how serious the situation is and how often it occurs in BNSL.

In BNSL, we seek its parent set $U$ of $X\in V$ s.t. the BD score $Q^n(X|U)$ is maximized among $U\subseteq S$
for each $S\subseteq V\backslash \{X\}$. The performance of BNSL depends on the prior probabilities over 
structures and over parameters. If we assume the former to be uniform, the solution of the decision
only depends on the choice of $a(\cdot)$. 

Let $c(u)$ and $c(x,u)$ be the occurrences of $U=u$ and $(U,X)=(u,x)$ in the $n$ examples
$(x_1,\cdots,x_n)$ and $(u_1,\cdots,u_n)$ that follow $X\in V$ and $U\subseteq V\backslash \{X\}$. 
We define the empirical entropy $H^n(X|U)$ of $X$ given $U$ by
$$H^n(X|U):= \sum_x\sum_u-\frac{c(x,u)}{n}\log\frac{c(x,u)}{c(u)}\ .$$
Empirical entropy expresses how the parent set $U$ fits to the $n$ examples for the variable $X$.

We define regularity of BNSL in order to proceed discussion.
\begin{teigi}\rm
BNSL is regular if $H^n(X|U)\leq H^n(X|U')$ and $U\subseteq U'$ imply $Q^n(X|U)\geq Q^n(X|U')$.
\end{teigi}
Regularity requires that  the fitness of $U$ to the $n$ examples and minimality of $U$ should be balanced.
For example, for information criteria such as AIC \cite{aic}, BIC \cite{bic}
$$AIC(X,U):=H^n(X|U)+\frac{k(U)}{n}\ {\rm and}\ BIC(X,U):=H^n(X|U)+\frac{k(U)}{2n}\log n$$
with $k(U)=(\alpha-1)\gamma$ satisfy
$$H^n(X|U)\leq H^n(X|U'), U\subseteq U' \Longrightarrow AIC(X,U)\leq AIC(X,U'),BIC(X,U)\leq BIC(X,U')\ ,$$ 
where $\alpha$ and $\gamma$ are the numbers of values that $X$ and $U$ take.
Furthermore, the BD score based on Jeffreys' prior 
$$-\log Q^n(X|U)=H^n(X|U)+\frac{k(U)}{2n}\log{n} +O(1)$$
in (\ref{eq412}) is regular.

Our main claim of this paper  is rather surprising:
\begin{teiri}\label{teiri2}\rm
BDSL based on the BDeu scores is not regular.
\end{teiri}
Proof. Let $Y\in V$ and $U\subseteq V\backslash \{X,Y\}$.
It is sufficient to show 
\begin{equation}\label{eq97}
Q^n(X|U)< Q^n(X|U\cup\{Y\})
\end{equation}
for some $(x_1,\cdots,x_n)$, $(y_1,\cdots,y_n)$, and $(u_1,\cdots,u_n)$ that follow $X,Y,U$ s.t. 
$H^n(X|U)\leq H^n(X|U\cup \{Y\})$.
Then, noting that (\ref{eq97}) can be written as 
$$\log Q^n(\{X,Y\}\cup U)+\log Q^n(U)-\log Q^n(\{X\}\cup U)-\log Q^n(\{Y\}\cup U)>0\ .$$
From Theorem 2 with ${\mathbf X}=\{X\}$, ${\mathbf Y}=\{Y\}$, and ${\mathbf Z}=U$,  if $f$ and $g$ exist s.t. $x_i=f(u_i)$ and $y_i=g(u_i)$ for $i=1,\cdots,n$, then (\ref{eq97}) follows.
We also observe $H^n(X|U)= H^n(X|U\cup \{Y\})=0$ when such $f$ and $g$ exist. This completes the proof.

\begin{rei}\rm \label{rei22}
Suppose that we have data in Table \ref{tab} ($n=12$) for four variables
$X,Y,Z,W$, and that we choose as the parent set of $X$ either $\{Z,W\}$ or $\{Z,W,Y\}$.
From the table, we notice that the values $x_i$ of $X$ are determined from
$u_i=(z_i,w_i)$ of $U$, so that the empirical entropy $H^n(X|U)$ is already zero.
Also, the information of $Y$ does not help for further dividing the states of $U$ because 
the values $y_i$ of $Y$ are determined from $u_i=(z_i,w_i)$ of $U$.
So, it is reasonable to stop adding more variables to $U=\{Z,W\}$.
However, from Theorem \ref{teiri2}, 
if we replace the current parent set candidate $U$ by $U\cup \{Y\}$,
the BDeu score (for any $\delta>0$) strictly increases 
because the values $y_i$ are determined from $u_i=(z_i,w_i)$ for $i=1,\cdots,n$.
In fact, for example, for the BDue score with $\delta=1$
$$Q^n(X|U)=\frac{Q^n(X,U)}{Q^n(U)}=\{\frac{\Gamma(3+1/8)}{\Gamma(1/8)}/\frac{\Gamma(3+1/4)}{\Gamma(1/4)}\}^3
=\{\frac{\frac{17}{8}\cdot \frac{9}{8}\cdot\frac{1}{8}}{\frac{9}{4}\cdot \frac{5}{4}\cdot\frac{1}{4}}\}^3=0.0767\cdots$$
$$Q^n(X|U\cup\{Y\})=\frac{Q^n(X,Y,Z,W)}{Q^n(Y,Z,W)}=\{\frac{\Gamma(3+1/16)}{\Gamma(1/16)}/\frac{\Gamma(3+1/8)}{\Gamma(1/8)}\}^3
=\{\frac{\frac{33}{16}\cdot \frac{17}{16}\cdot\frac{1}{16}}{\frac{17}{8}\cdot \frac{9}{8}\cdot\frac{1}{8}}\}^3=0.0962\cdots\ ,$$
so that $Q^n(X|U)<Q^n(X|U\cup\{Y\})$. Thus, BDeu chooses $\{Y,Z,W\}$ rather than $\{Z,W\}$ as the parent set,
which demonstrates that BNSL based on the BDeu scores is not regular.
\end{rei}

\begin{table}
\caption{Data of Variables $X$, $U=\{Z,W\}$, and $Y$ ($n=12$)  \label{tab}}
\begin{center}
\begin{tabular}{l|llllllllllll}
\hline
$X$&0&0&0&1&1&1&1&1&1&0&0&0\\
\hline
$Z$&0&0&0&0&0&0&1&1&1&1&1&1\\ 
$W$&0&0&0&1&1&1&0&0&0&1&1&1\\
\hline
$Y$&0&0&0&0&0&0&0&0&0&1&1&1\\
\hline
\end{tabular}
\end{center}
\end{table}

\section{Concluding Remarks}
In this paper, we defined regularity of BNSL, i.e.,
$$H^n(X|U)\leq H^n(X|U'), U\subseteq U' \Longrightarrow Q^n(X|U)\geq Q^n(X|U')$$
when we seek a parent set of $X$ and choose either $U$ or $U'$,
and proved that the BDeu violates the condition. In order to see why the phenomenon occurs,
we defined the quantity $J(n)$ and proved that the probability of $J(n)>0$ converges to zero as $n\rightarrow \infty$
pointwise for the BDeu scores and uniformly for the BD scores based on the Jeffreys' prior, when ${\mathbf X}\ci {\mathbf Y}|{\mathbf Z}$.

At the same time, we demonstrated that this phenomenon causes serious problems in BNSL.
Even when the state space has been decomposed and the empirical entropy $H^n(X|U)$ 
has reached to zero, the BDeu still further seeks a more refined state space by choosing additional variables.
This suggests that in the BDeu scores,
the fitness of a BN structure to the examples and simplicity of the BN structure are not balanced, 
which should be avoided in any model selection procedure.

Although through the paper, we might have stresses the demerit of the BDeu more than necessary,
it may be possible to find more merits than the demerits such as the conditional score can be expressed as in 
(\ref{eq155}). However, nothing has been proved for any property w.r.t. estimation.
It seems that the almost all existing researches on this topic claim their merits only through experiments
from which no persuasive conclusion can be obtained.
On the other hand, this paper mathematically derived important properties of the BDeu.

There is an NML-based score for BNSL (Silander et al. 2008 \cite{skm2}) suggested for replacing BDeu. 
Adding it to the comparison would be included in our future work.

\section*{Appendix A: Proof of Theorem 1}
We only prove the simplest case ${\mathbf X}=\{X\}$, ${\mathbf Y}=\{Y\}$, and ${\mathbf Z}=\{\}$.
The general case can be obtained in a straightforward manner. 

Using Stirling's formula, 
$$\log \Gamma(z)=z\log z-z +\frac{1}{2}\log \frac{2\pi}{z}+\epsilon(z)$$
with $\displaystyle \frac{1}{12z}<\epsilon(z)<\frac{1}{12z+1}$, we have
\begin{eqnarray*}
&&-\log Q^n(X)\\
&=&\log\Gamma(\sum_x\{c(x)+a(x)\})-\log\Gamma(\sum_x a(x))-\sum_x\log\Gamma(c(x)+a(x))+\sum_x\log\Gamma(a(x))\\
&=&-\sum_x \{c(x)+a(x)\}\log\frac{c(x)+a(x)}{\sum_{x'} \{c(x')+a(x')\}}
+\frac{1}{2}\sum_x\log\{c(x)+a(x)\}\\
&&-\frac{1}{2}\log {\sum_x\{c(x)+a(x)\}}-\frac{\alpha-1}{2}\log 2\pi+\epsilon(\sum_z\{c(x)+a(x)\})-\sum_z\epsilon(c(x)+a(x))\ ,
\end{eqnarray*}
where the last three terms are $O(1)$

For the BD score based on Jeffreys' prior, 
we assume $a(z)=0.5$, thus we have
$$-\log Q^n(Z) = \sum_zc(z)\log\frac{n+\gamma/2}{c(z)+1/2}+\frac{\gamma-1}{2}\log {n}+O(1)$$
From
$$-\frac{\gamma}{2}\leq \sum_zc(z)\log \frac{1+\frac{\gamma}{2n}}{1+\frac{1}{2c(z)}}\leq 0 \ ,$$
we have 
$$-\log Q^n(Z) = \sum_zc(z)\log\frac{n}{c(z)}+\frac{\gamma-1}{2}\log {n}+O(1)\ .$$
Similarly, we have
$$-\log Q^n(X,Z) = \sum_x\sum_zc(x,z)\log\frac{n}{c(x,z)}+\frac{\alpha\gamma-1}{2}\log {n}+O(1)\ .$$
$$-\log Q^n(Y,Z) = \sum_y\sum_zc(y,z)\log\frac{n}{c(y,z)}+\frac{\beta\gamma-1}{2}\log {n}+O(1)\ .$$
and
$$-\log Q^n(X,Y,Z) = \sum_x\sum_y\sum_zc(x,y,z)\log\frac{n}{c(x,y,z)}+\frac{\alpha\beta\gamma-1}{2}
\log{n}+O(1)\ .$$
Thus, we have (\ref{eq412}).

For the BDeu score, on the other hand, 
we assume $a(z)=\delta/\gamma$, 
thus we have
$$-\log Q^n(Z)= 
\sum_zc(z)\log\frac{n+\delta}{c(z)+\frac{\delta}{\gamma}}+(\delta-\frac{1}{2})\log (n+\delta)-(\frac{\delta}{\gamma}-\frac{1}{2})\sum_z\log 
\{c(z)+\frac{\delta}{\gamma}\}+O(1)
$$
From
$$-\frac{\delta}{\gamma}\leq \sum_x c(z)\log \frac{1+\frac{\delta}{n}}{1+\frac{\delta}{\gamma c(z)}}\leq 0\ ,$$
we have
$$
-\log Q^n(Z)=\sum_zc(z)\log\frac{n}{c(z)}+\frac{\gamma-1}{2}\log{n}-(\frac{\delta}{\gamma}-\frac{1}{2})\sum_z\log \frac{c(z)+\frac{\delta}{\gamma}}{n+\delta}
+O(1)\ .
$$
Similarly, we have
\begin{eqnarray*}
-\log Q^n(X,Z)
=\sum_x\sum_zc(x,z)\log\frac{n}{c(x,z)}+\frac{\alpha\beta-1}{2}\log{n}
-(\frac{\delta}{\alpha\gamma}
-\frac{1}{2})\sum_x\sum_z\log \frac{c(x,z)+\frac{\delta}{\alpha\gamma}}{n+\delta}+O(1)\ .
\end{eqnarray*}
\begin{eqnarray*}
-\log Q^n(Y,Z)
=\sum_y\sum_zc(y,z)\log\frac{n}{c(y,z)}+\frac{\beta\gamma-1}{2}\log{n}
-(\frac{\delta}{\beta\gamma}
-\frac{1}{2})\sum_y\sum_z\log \frac{c(y,z)+\frac{\delta}{\beta\gamma}}{n+\delta}+O(1)\ .
\end{eqnarray*}
and
\begin{eqnarray*}
-\log Q^n(X,Y,Z)&=&\sum_x\sum_y\sum_zc(x,y,z)\log\frac{n}{c(x,y,z)}+\frac{\alpha\beta\gamma-1}{2}\log{n}\\
&&-(\frac{\delta}{\alpha\beta\gamma}
-\frac{1}{2})\sum_x\sum_y\sum_z\log \frac{c(x,y,z)+\frac{\delta}{\alpha\beta\gamma}}{n+\delta}+O(1)\ .
\end{eqnarray*}
Thus, we have (12).

\section*{Appendix B: Proof of Theorem 2}
We only prove the simplest case ${\mathbf X}=\{X\}$, ${\mathbf Y}=\{Y\}$, and ${\mathbf Z}=\{\}$.
The general case can be obtained in a straightforward manner. 

For Jeffreys', 
we show $Q^n(X)Q^n(Y)\geq Q^n(XY)$ for $n\geq 1$.
Since $c_X(0)=c_Y(0)=n$ and $c_{XY}(0.0)=n$ in
$$
\frac{\Gamma(\alpha/2)}{\Gamma(n+\alpha/2)}\prod_x\frac{\Gamma(c_X(x)+1/2)}{\Gamma(1/2)}
\cdot
\frac{\Gamma(\beta/2)}{\Gamma(n+\beta/2)}\prod_y\frac{\Gamma(c_Y(y)+1/2)}{\Gamma(1/2)}
$$
and
$$
\frac{\Gamma(\alpha\beta/2)}{\Gamma(n+\alpha\beta/2)}\prod_x\prod_y\frac{\Gamma(c_{XY}(x,y)+1/2)}{\Gamma(1/2)}\ ,
$$
it is sufficient to show 
$$
\frac{\Gamma(\alpha/2)}{\Gamma(n+\alpha/2)}\frac{\Gamma(n+1/2)}{\Gamma(1/2)}
\cdot
\frac{\Gamma(\beta/2)}{\Gamma(n+\beta/2)}\frac{\Gamma(n+1/2)}{\Gamma(1/2)}\geq 
\frac{\Gamma(\alpha\beta/2)}{\Gamma(n+\alpha\beta/2)}\frac{\Gamma(n+1/2)}{\Gamma(1/2)}\ ,
$$
which means 
\begin{equation}\label{eq981}
\frac{\Gamma(n+\alpha\beta/2)\Gamma(n+1/2)}{\Gamma(\alpha\beta/2)\Gamma(1/2)}\geq
 \frac{\Gamma(n+\alpha/2)\Gamma(n+\beta/2)}{\Gamma(\alpha/2)\Gamma(\beta/2)}.
\end{equation}
For $n=0$, the both sides are 1, and the inequality (\ref{eq981}) is true.
Then, we find that (\ref{eq981}) for $n$ implies (\ref{eq981}) for $n+1$ because we see
\begin{eqnarray*}
&&\Gamma(n+1+\alpha/2)\Gamma(n+1+\alpha/2)=\Gamma(n+\alpha/2)\Gamma(n+\alpha/2)\cdot (n+\alpha/2)(n+\beta/2)\\
&\leq& \Gamma(n+\alpha\beta/2)\Gamma(n+\frac{1}{2})\cdot (n+\alpha/2)(n+\beta/2)\\
&\leq & \Gamma(n+\alpha\beta/2)\Gamma(n+1/2) \cdot (n^2+(\alpha\beta+1)n/2+\alpha\beta/4)\\
&=&\Gamma(n+\alpha\beta/2)(n+1/2)\cdot (n+{\alpha\beta}/{2})(n+1/2)=\Gamma(n+1+\alpha\beta/2)(n+1+1/2)\ ,
\end{eqnarray*}
where the first inequality follows from the assumption of induction.

For BDeu with equivalent sample size $\delta>0$, we show we show $Q^n(X)Q^n(Y)\leq Q^n(XY)$ for $n\geq 1$. Then, the both sides will be replaced by
$$
\frac{\Gamma(\delta)}{\Gamma(n+\delta)}\prod_x\frac{\Gamma(c_X(x)+\delta/\alpha)}{\Gamma(\delta/\alpha)}
\cdot
\frac{\Gamma(\delta)}{\Gamma(n+\delta)}\prod_y\frac{\Gamma(c_Y(y)+\delta/\beta)}{\Gamma(\delta/\beta)}
$$
and
$$
\frac{\Gamma(\delta)}{\Gamma(n+\delta)}\prod_x\prod_y\frac{\Gamma(c_{XY}(x,y)+\delta/\alpha\beta)}{\Gamma(\delta/\alpha\beta)}\ ,
$$
and it is sufficient to show 
\begin{equation}\label{eq982}
\frac{\Gamma(n+\delta/\alpha)}{\Gamma(\delta/\alpha)}\cdot \frac{\Gamma(n+\delta/\alpha)}{\Gamma(\delta/\alpha)}
 \leq \frac{\Gamma(n+\delta/\alpha\beta)}{\Gamma(\delta/\alpha\beta)}\cdot \frac{\Gamma(n+\delta)}{\Gamma(\delta)}\ .
\end{equation}
Then, we find that (\ref{eq982}) for $n$ implies (\ref{eq982}) for $n+1$ by induction and 
$(n+\delta/\alpha)(n+\delta/\beta)\leq (n+\delta)(n+\delta/\alpha\beta)$.

In particular, the equality does not hold for $n=2$, so that a strict inequality holds for $n\geq 2$.

This completes the proof.

\bibliography{2016-11-20}

\end{document}